\documentclass{article}

\usepackage[preprint]{neurips_2026}

\usepackage[utf8]{inputenc}
\usepackage[T1]{fontenc}
\usepackage{hyperref}
\usepackage{url}
\usepackage{booktabs}
\usepackage{amsfonts}
\usepackage{amsmath}
\usepackage{nicefrac}
\usepackage{microtype}
\usepackage{xcolor}
\usepackage{graphicx}
\usepackage{multirow}
\usepackage{threeparttable}
\usepackage{float}
\usepackage{enumitem}
\usepackage{subcaption}
\usepackage{wrapfig}

\newcommand{\boldres}[1]{\textcolor{red}{\textbf{#1}}}
\newcommand{\secondres}[1]{\textcolor{blue}{\underline{#1}}}

\title{DecompKAN: Decomposed Patch-KAN for Long-Term Time Series Forecasting}

\author{Naveen Mysore \\
\texttt{nmysore.work@gmail.com}}

\begin{document}
\maketitle

\begin{abstract}

Accurate time series forecasting in scientific domains such as climate modeling, physiological monitoring, and energy systems benefits from both competitive predictions and model transparency: practitioners value understanding \emph{how} a model transforms temporal features, not merely \emph{what} it predicts. Transformer-based models achieve strong accuracy but their attention weights reveal only token-level relevance, not the functional transformations applied to each feature. This work proposes \textsc{DecompKAN}, a lightweight attention-free architecture that combines trend--residual decomposition, channel-wise patching, learned instance normalization, and B-spline Kolmogorov--Arnold Network (KAN) edge functions. Each KAN edge learns an explicit, inspectable 1D scalar function $\phi(x)$ over learned patch-embedding coordinates that can be directly visualized, offering a form of architectural transparency not directly available in attention-based or MLP-based architectures. On standard benchmarks, \textsc{DecompKAN} achieves best or tied-best MSE on 15 of 32 dataset--horizon combinations among selected published baselines, and achieves best or tied-best MSE on 20 of 36 comparisons (25 of 36 MAE; ties counted for all tied models) under a controlled same-recipe evaluation across 9 datasets including the physiological PPG-DaLiA benchmark. The architecture shows particular strength on datasets with smooth temporal dynamics (Solar~$-17\%$, ECL~$-10\%$ vs.\ iTransformer, Weather) and physiological time series (PPG-DaLiA). Visualization of learned edge functions reveals qualitatively different latent nonlinearities across domains: oscillatory patterns for meteorological data and threshold-like activations for physiological signals. Ablation analysis shows that the architectural pipeline (decomposition, patching, normalization) drives performance more than the choice of nonlinear layer, while the KAN formulation enables inspection of learned latent transformations.

\end{abstract}

\section{Introduction}
\enlargethispage{3\baselineskip}

\begin{wrapfigure}{r}{0.38\textwidth}
\centering
\vspace{-20pt}
\includegraphics[width=0.36\textwidth]{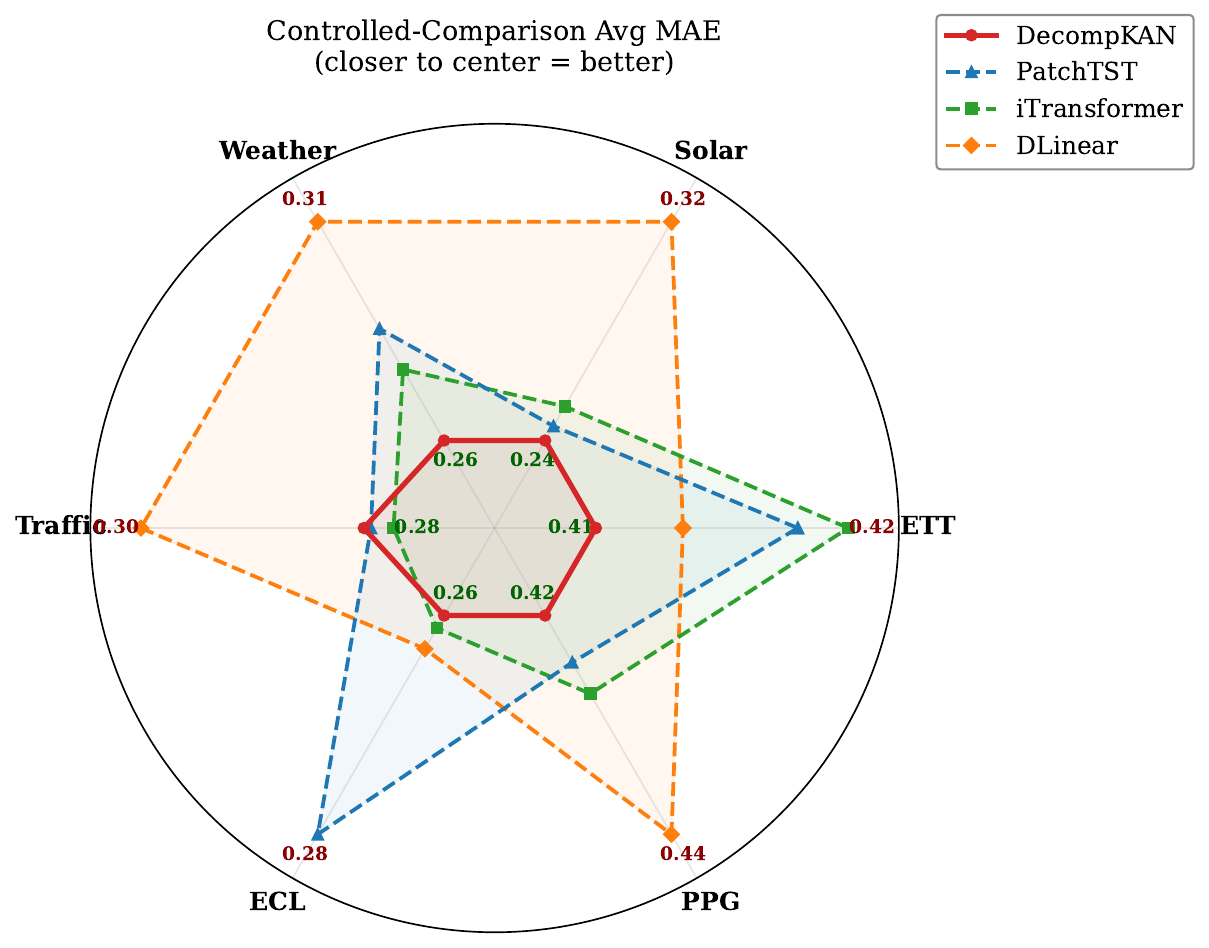}
\vspace{-10pt}
\caption{Controlled-comparison avg MAE (lower = better). ETT = ETTh1/h2/m1/m2. \textsc{DecompKAN} ranks first on 25/36 (Table~\ref{tab:controlled}).}
\label{fig:radar}
\vspace{-15pt}
\end{wrapfigure}

Long-term time series forecasting (LTSF) is a cornerstone of decision-making in climate science, physiological monitoring, energy systems, and finance. In scientific applications, forecasting serves a dual purpose: producing accurate predictions \emph{and} providing mechanistic insight into the temporal dynamics that generate those predictions. A climate scientist forecasting atmospheric pressure needs to verify that the model captures known physical relationships (temperature--humidity coupling, diurnal cycles); a clinician monitoring ICU vital signs needs to understand whether a heart rate forecast reflects learned physiological dynamics or spurious correlations.

The dominant forecasting paradigm, transformer-based models with self-attention~\citep{vaswani2017attention}, achieves strong accuracy but offers limited mechanistic transparency. Attention weights reveal \emph{which} input tokens influence the output, but not \emph{what functional transformation} is applied to each temporal feature. Post-hoc methods (gradient saliency, SHAP) provide local approximations but do not faithfully expose the model's internal computation. For scientific domains where predictions must be explainable in terms of learned input--output relationships, this opacity is a fundamental limitation.

This work investigates two questions: (1)~whether a carefully designed feedforward pipeline of decomposition, patching, and normalization can achieve competitive forecasting without attention, and (2)~whether replacing the feedforward core with B-spline KAN edge functions~\citep{liu2024kan} provides useful inspectability. Each KAN edge learns a 1D nonlinear scalar function $\phi_{i \to j}(x)$ over learned patch-embedding coordinates that can be directly visualized; mapping these latent functions back to raw variables requires additional attribution analysis, but the functional form itself is available for inspection.

The hypothesis is instantiated in \textsc{DecompKAN}, which combines three components:
\begin{enumerate}[nosep]
    \item \textbf{Decomposed Patch-KAN.} Moving-average decomposition separates the input into trend and residual components, each processed by an independent channel-wise Patch-KAN branch.
    \item \textbf{Learned Instance Normalization.} A stacked RevIN + adaptive normalization module handles non-stationarity without manual per-dataset configuration. The fixed RevIN layer removes window-level drift; the learned adaptive layer fine-tunes the normalization per dataset.
    \item \textbf{B-spline edge functions with intrinsic interpretability.} Each KAN edge learns a smooth 1D scalar function via cubic B-splines with local support, providing stable optimization and smooth interpolation. Unlike attention weights (which show \emph{where} a model looks) or gradient saliency maps (which are post-hoc approximations), each edge function $\phi_{i \to j}(x)$ is a directly inspectable 1D transformation over learned patch-embedding coordinates. This enables inspection of what nonlinear operations the model applies, which is valuable in domains such as clinical decision support, atmospheric science, and energy forecasting.
\end{enumerate}

The contributions of this work are:
\begin{itemize}[nosep]
    \item \textbf{Inspectable latent edge functions.} Each KAN edge function $\phi_{i \to j}(x)$ is a directly inspectable 1D nonlinear transformation over learned patch-embedding coordinates. Visualization on Weather and PPG-DaLiA (Appendix~\ref{app:interpretability}) reveals qualitatively different latent nonlinearities across domains (oscillatory for meteorological data, threshold-like for physiological signals) and structured sparsity. Mapping these latent functions back to raw variables requires additional attribution analysis, but the functional form is directly available for inspection.
    \item \textbf{Competitive accuracy with particular strength on physics-structured data.} Best or tied-best MSE on 15/32 dataset--horizon combinations among selected published baselines, and achieves best or tied-best MSE on 20 of 36 comparisons (25 of 36 MAE; ties counted for all tied models) under a controlled same-recipe evaluation across 9 datasets including PPG-DaLiA. Particular strength on Solar ($-17\%$), ECL ($-10\%$ vs.\ iTransformer), Weather, and physiological data, using 1.9--3.6M parameters (Appendix~\ref{tab:params}).
    \item \textbf{Pipeline design matters more than component choice.} Ablation shows decomposition and RevIN are the largest contributors (up to $+$4.8\% degradation when removed); replacing KAN with linear or attention layers yields comparable results.
    \item \textbf{Controlled comparison.} Retraining baselines with \textsc{DecompKAN}'s protocol illustrates sensitivity of relative rankings to training recipe.
\end{itemize}

\section{Related Work}

\textbf{Transformers for LTSF.}\quad Informer~\citep{zhou2021informer} introduced sparse attention for long sequences. Autoformer~\citep{wu2021autoformer} embedded series decomposition with auto-correlation. FEDformer~\citep{zhou2022fedformer} operated in the frequency domain. DLinear~\citep{zeng2023dlinear} challenged all of these with decomposition + linear layers, establishing that architectural choices (patching, normalization, decomposition) matter more than attention itself. PatchTST~\citep{nie2023patchtst} reconciled transformers with patching and channel independence. iTransformer~\citep{liu2024itransformer} inverted attention across variates for high-dimensional datasets. TimesNet~\citep{wu2023timesnet} modeled temporal 2D variations. CATS~\citep{lu2024cats} constructs auxiliary time series as exogenous variables to capture inter-series relationships with a simple MLP predictor, achieving strong results particularly on ETT datasets. The present work retains patching and channel independence from PatchTST but replaces attention entirely with KAN, while still outperforming iTransformer on ECL (321 variates, $-10\%$).

\textbf{KAN for time series.}\quad The Kolmogorov--Arnold representation theorem~\citep{kolmogorov1957representation}, as adapted into neural-network form by \citet{liu2024kan}, motivates replacing fixed MLP activations with learnable B-spline edge functions. RMoK~\citep{xu2024kan4tsf} combines B-spline, wavelet, Taylor, and Jacobi KAN variants via mixture-of-experts with RevIN, but lacks decomposition and patching. TimeKAN~\citep{qiu2024timekan} uses Chebyshev KAN with frequency decomposition (12--38K params), excelling on ETT datasets. Both outperform \textsc{DecompKAN} on datasets with complex regime changes (ETTh1, ETTm1), while \textsc{DecompKAN} is stronger on physics-structured datasets (Weather, Solar, ECL). SimpleTM~\citep{chen2025simpletm} explores simple temporal mixtures, confirming the viability of lightweight architectures.

\textbf{Foundation models for time series.}\quad A recent line of work applies large pre-trained models to time series forecasting. Moirai~\citep{woo2024moirai} trains a masked encoder on diverse time series corpora and achieves competitive zero-shot forecasting. Chronos~\citep{ansari2024chronos} tokenizes time series values and fine-tunes a T5-based language model. TimesFM~\citep{das2024timesfm} pre-trains a decoder-only transformer on large-scale temporal data. TimeLLM~\citep{jin2024timellm} reprograms frozen LLMs for forecasting via prompt-based input reprogramming. These foundation models leverage broad pre-training but require significant compute and data. In contrast, \textsc{DecompKAN} is a lightweight trained-from-scratch architecture that achieves competitive results on several benchmarks without large-scale pre-training.

\textbf{Normalization for non-stationarity.}\quad RevIN~\citep{kim2022revin} normalizes per-window statistics but is not always optimal. In the ablation presented in this work (Table~\ref{tab:ablation}), removing RevIN causes only modest degradation on some datasets. Non-stationary Transformers~\citep{liu2022nonstationary} and Dish-TS~\citep{fan2023dishts} address distribution shift through de-stationary attention and learned dish initialization. The present work stacks RevIN with a learned adaptive layer whose denormalization is \emph{separately parameterized} (not the mathematical inverse of normalization), enabling per-dataset adaptation without manual toggles. The adaptive module receives only the RevIN-normalized sequence (not the original $\mu_c, \sigma_c$) and learns scale/shift adjustments based on the temporal shape of the normalized signal; RevIN's inverse separately restores the original scale at output time.

\section{Methodology and Experiments}

This section presents the \textsc{DecompKAN} architecture, followed by the training protocol and evaluation setup.

\subsection{Architecture}

\begin{figure}[t]
\centering
\includegraphics[width=\textwidth]{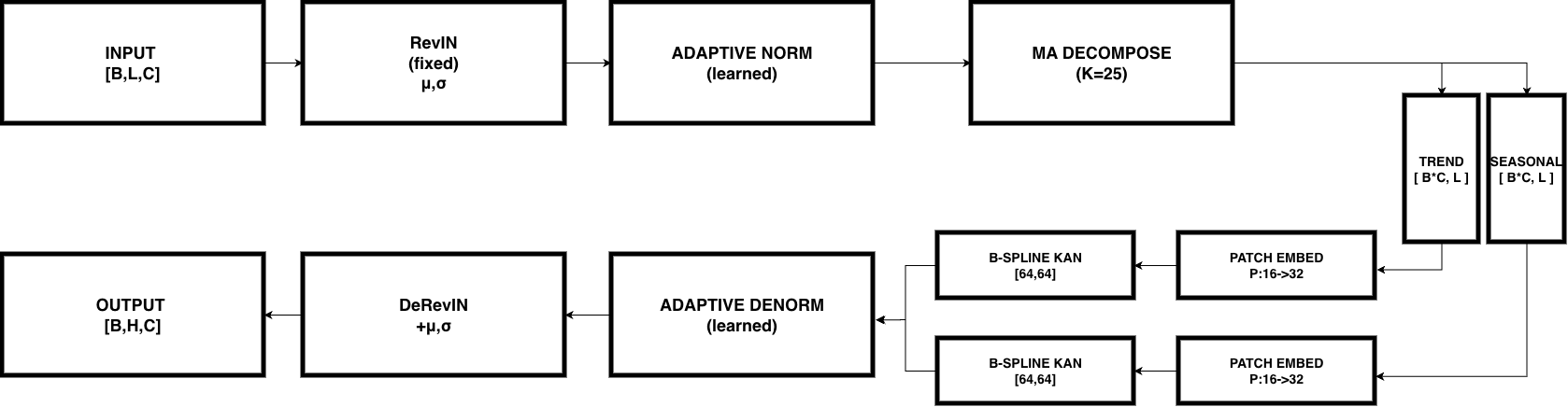}
\caption{\textsc{DecompKAN} architecture. Input is normalized (RevIN + learned adaptive), decomposed via moving average ($K_{\mathrm{MA}}$=25), and each component is forecast by an independent Patch-KAN branch. Outputs are summed and denormalized. Parameter count varies with $L$ and $H$ (see Table~\ref{tab:params}).}
\label{fig:architecture}
\end{figure}

\textsc{DecompKAN} processes $\mathbf{x} \in \mathbb{R}^{L \times C}$ to forecast $\hat{\mathbf{y}} \in \mathbb{R}^{H \times C}$ through four channel-independent stages (Figure~\ref{fig:architecture}).

\textbf{Stacked normalization.}\quad The first stage applies RevIN~\citep{kim2022revin} to remove per-window drift: $\tilde{\mathbf{x}}_c = (\mathbf{x}_c - \mu_c) / \sigma_c$ where $\mu_c$ and $\sigma_c$ are the per-channel mean and standard deviation of the input window. An adaptive module then applies a learned scale and shift:
\begin{equation}
\hat{\mathbf{x}}_c = s_c \cdot \tilde{\mathbf{x}}_c + b_c, \quad (s_c, b_c) = g_\theta(\tilde{\mathbf{x}}_c)
\end{equation}
where $g_\theta: \mathbb{R}^L \rightarrow \mathbb{R}^2$ is a small MLP (Linear($L$, 32) $\rightarrow$ GELU $\rightarrow$ Linear(32, $d_s$) $\rightarrow$ Linear($d_s$, 2) with $d_s{=}8$) that maps the full $L$-length RevIN-normalized sequence per channel into a learned scale $s_c$ and shift $b_c$. Critically, the denormalization head is \emph{separately parameterized}, not the mathematical inverse of the normalization, enabling asymmetric transforms. Both heads are initialized as identity transforms. Note that the adaptive module receives the RevIN-normalized signal (mean${\approx}0$, std${\approx}1$), so it does not have direct access to the absolute scale removed by RevIN. Instead, it learns dataset-specific adjustments based on higher-order distributional features (temporal shape, skewness, local structure) of the normalized signal. RevIN's inverse transform separately restores the original scale at output time.

\textbf{Trend--residual decomposition.}\quad Following the decomposition design of \citet{wu2021autoformer} and \citet{zeng2023dlinear}, a moving-average filter with kernel size $K_{\mathrm{MA}}{=}25$ separates the trend component $\mathbf{s}_\tau$ from the residual component $\mathbf{s}_\sigma = \hat{\mathbf{x}} - \mathbf{s}_\tau$. Decomposition separates slow trend from fast oscillation before patch embedding. Note that the KAN receives learned patch embeddings, not raw decomposed values, so the precise mechanism by which decomposition helps is not fully clear; the patch embedding can rescale and mix values before the B-spline functions see them. Empirically, decomposition consistently improves performance (Table~\ref{tab:ablation}), but whether this benefit is specific to KAN is not established by the current ablation. Preliminary synthetic experiments (Appendix~\ref{app:synthetic}) suggest that decomposition combined with KAN can substantially outperform either approach alone on non-stationary signals.

\textbf{Patch-KAN branches.}\quad Following the patching strategy of \citet{nie2023patchtst}, each decomposed component is processed by an independent Patch-KAN branch. The input sequence of length $L$ is divided into $N = \lfloor(L - P)/S\rfloor + 1$ overlapping patches of length $P{=}16$ with stride $S{=}8$. Each patch is linearly embedded into dimension $d{=}32$, producing a representation of size $N \times d$. This is flattened and processed by a KAN with three transformation layers (two hidden layers of width 64, one output layer of width $H$):
\begin{equation}
\text{KAN}: \mathbb{R}^{Nd} \xrightarrow{\text{layer 1}} \mathbb{R}^{64} \xrightarrow{\text{layer 2}} \mathbb{R}^{64} \xrightarrow{\text{layer 3}} \mathbb{R}^{H}
\end{equation}
Following \citet{liu2024kan}, each KAN edge computes $\phi(x) = w \cdot \text{SiLU}(x) + \sum_{i} c_i B_i^p(x)$, where $B_i^p$ are cubic ($p{=}3$) B-spline basis functions~\citep{deboor1978splines} with local support on a uniform grid of size $G{=}5$, $c_i$ are learnable coefficients, and $w$ is a residual linear weight. The SiLU residual path ensures gradient flow even when spline coefficients are poorly initialized. The trend and residual branch outputs are summed and reshaped to $\mathbb{R}^{H \times C}$.

\textbf{Denormalization.}\quad The output passes through an adaptive denormalization head, then through RevIN's inverse transform ($\hat{\mathbf{y}}_c = \hat{\mathbf{y}}'_c \cdot \sigma_c + \mu_c$). The denormalization head reuses the same statistics vector computed from the \emph{input} sequence during normalization (not from the forecast output), but applies a separately parameterized MLP ($d_s \rightarrow 8 \rightarrow 2$) to produce output-specific scale and shift. This asymmetry between normalization and denormalization paths gives the model flexibility to learn different input-to-output scale relationships per dataset.

\textbf{Channel independence.}\quad Following \citet{nie2023patchtst}, all operations are shared across the $C$ variates (channel-independent design). Each variate is processed by the same weights, and the batch dimension is expanded to $B \times C$ before the KAN branches. This provides implicit regularization. Channel independence was adopted after preliminary (non-exhaustive) experiments with cross-variate alternatives did not consistently improve performance (Section~\ref{sec:cross_variate}).

\subsection{Training Protocol}
\label{sec:training}

All models are trained with MSE loss, Adam~\citep{kingma2015adam} optimizer, cosine learning rate schedule with linear warmup (10\% of total steps), and gradient clipping at norm 1.0. Batch size is 32, maximum epochs 50 with early stopping (patience 10). Per-dataset hyperparameters are selected from a grid of 12 configurations (3 learning rates $\times$ 2 bidirectional settings $\times$ 2 lookback lengths) validated at $H{=}96$:
\begin{itemize}[nosep]
    \item Learning rate $\in \{10^{-3},\ 5{\times}10^{-4},\ 2{\times}10^{-4}\}$
    \item Bidirectional augmentation $\in \{\text{on}, \text{off}\}$ (time-reversed windows mixed with forward)
    \item Lookback length $L \in \{336, 512\}$
\end{itemize}
All other hyperparameters ($P{=}16$, $S{=}8$, $d{=}32$, KAN hidden$={64}$, depth$={2}$, grid size $G{=}5$, spline order $p{=}3$, MA kernel $K_{\mathrm{MA}}{=}25$) are fixed across all datasets. The best configuration per dataset is reported in Appendix~\ref{app:tuning}.

\textbf{Bidirectional augmentation.}\quad For applicable datasets, training data is augmented by reversing each window in time and training the model to predict the reverse-direction horizon. This doubles the effective training set and acts as an empirical regularizer. The augmentation is applied only during training; evaluation uses standard forward-time windows. Per-dataset tuning determines whether it is beneficial (Appendix~\ref{app:tuning}). An ablation of this component is reported in Appendix~\ref{app:ablation}: removing it degrades ETTh1 by $+$10.0\% but affects Weather by only $+$0.6\%, indicating that its contribution is dataset-dependent.

\subsection{Evaluation Setup}

\textbf{Datasets.}\quad Eight standard LTSF benchmarks are used for the published comparison: Weather~\citep{wu2021autoformer} (21 variates, meteorological), Solar, ECL, and Traffic~\citep{lai2018lstnet} (137, 321, and 862 variates respectively), and ETTh1/h2/m1/m2~\citep{zhou2021informer} (7 variates, hourly or 15-minute transformer data), yielding 32 dataset--horizon combinations. The controlled comparison additionally includes PPG-DaLiA~\citep{reiss2019ppg}, a multi-sensor physiological dataset from 15 subjects. We extract 15 sensor channels: chest ECG, EMG, EDA, temperature, respiration, and 3-axis acceleration (from RespiBAN at 700Hz), plus wrist BVP, EDA, temperature, and 3-axis acceleration (from Empatica E4 at 4--64Hz), plus interpolated heart-rate labels. All channels are resampled to 4Hz via decimation, then subsampled to 1Hz. Subjects are concatenated chronologically (not subject-disjoint) with a standard 70/10/20 split, yielding 52{,}696 timesteps and 36 controlled comparisons. Forecast horizons are $H \in \{96, 192, 336, 720\}$ for all datasets.

\textbf{Baselines.}\quad Two comparison modes are employed:
\begin{enumerate}[nosep]
    \item \emph{Published comparison} (Table~\ref{tab:main}): Results from the original papers of iTransformer~\citep{liu2024itransformer}, PatchTST~\citep{nie2023patchtst}, DLinear~\citep{zeng2023dlinear}, TimeKAN~\citep{qiu2024timekan}, and RMoK~\citep{xu2024kan4tsf}.
    \item \emph{Controlled comparison} (Table~\ref{tab:controlled}): PatchTST, iTransformer, and DLinear retrained with the \textsc{DecompKAN} training protocol (same $L$, batch size, epochs, LR schedule, augmentation, gradient clipping) to assess protocol sensitivity under a common recipe. These results should not be interpreted as architecture-isolated rankings, since the recipe was developed for \textsc{DecompKAN}.
\end{enumerate}

\textbf{Metrics.}\quad Mean Squared Error (MSE) and Mean Absolute Error (MAE) on the test set, following standard protocol.

\section{Results}

\subsection{Main Results}

\begin{table}[!htbp]
\caption{Multivariate long-term forecasting (MSE/MAE). $H \in \{96, 192, 336, 720\}$. \boldres{Red}: best. \secondres{Blue}: second. ``--'': not reported. \textsc{DecompKAN} results for Weather, ETTh1/h2, ETTm2 are 10-seed means (Table~\ref{tab:seeds}); others are single seed. Baselines from original papers: iTransformer~\citep{liu2024itransformer} Table~10; PatchTST~\citep{nie2023patchtst} Table~3; DLinear~\citep{zeng2023dlinear} Table~2; RMoK-B~\citep{xu2024kan4tsf} Table~2; TimeKAN~\citep{qiu2024timekan}. First counts use unrounded values; rounded ties may not reproduce counts exactly.}
\label{tab:main}
\vskip 0.02in
\centering
\resizebox{\columnwidth}{!}{
\begin{threeparttable}
\begin{scriptsize}
\renewcommand{\arraystretch}{0.85}
\setlength{\tabcolsep}{2.5pt}
\begin{tabular}{c|c|cc|cc|cc|cc|cc|cc}
\toprule
\multicolumn{2}{c}{\multirow{2}{*}{Models}} &
\multicolumn{2}{c}{\textbf{DecompKAN}} &
\multicolumn{2}{c}{TimeKAN} &
\multicolumn{2}{c}{RMoK-B} &
\multicolumn{2}{c}{iTransformer} &
\multicolumn{2}{c}{PatchTST} &
\multicolumn{2}{c}{DLinear} \\
\multicolumn{2}{c}{} &
\multicolumn{2}{c}{\textbf{(This work)}} &
\multicolumn{2}{c}{\scalebox{0.9}{\citeyear{qiu2024timekan}}} &
\multicolumn{2}{c}{\scalebox{0.9}{\citeyear{xu2024kan4tsf}}} &
\multicolumn{2}{c}{\scalebox{0.9}{\citeyear{liu2024itransformer}}} &
\multicolumn{2}{c}{\scalebox{0.9}{\citeyear{nie2023patchtst}}} &
\multicolumn{2}{c}{\scalebox{0.9}{\citeyear{zeng2023dlinear}}} \\
\cmidrule(lr){3-4}\cmidrule(lr){5-6}\cmidrule(lr){7-8}\cmidrule(lr){9-10}\cmidrule(lr){11-12}\cmidrule(lr){13-14}
\multicolumn{2}{c}{Metric} & MSE & MAE & MSE & MAE & MSE & MAE & MSE & MAE & MSE & MAE & MSE & MAE \\
\toprule
\multirow{5}{*}{\rotatebox{90}{Weather}}
 &96 & \boldres{.148} & \boldres{.195} & .162 & .208 & .171 & .221 & .174 & .214 & \secondres{.149} & \secondres{.198} & .176 & .237 \\
 &192 & \boldres{.192} & \boldres{.238} & .207 & .249 & .220 & .263 & .221 & .254 & \secondres{.194} & \secondres{.241} & .220 & .282 \\
 &336 & \boldres{.243} & \boldres{.277} & .263 & .290 & .277 & .302 & .278 & .296 & \secondres{.245} & \secondres{.282} & .265 & .319 \\
 &720 & \secondres{.321} & \boldres{.331} & .338 & .340 & .360 & .354 & .358 & .347 & \boldres{.314} & \secondres{.334} & .323 & .362 \\
\cmidrule(lr){2-14}
 &Avg & \boldres{.226} & \boldres{.260} & .243 & .272 & .257 & .285 & .258 & .278 & \secondres{.226} & \secondres{.264} & .246 & .300 \\
\midrule
\multirow{5}{*}{\rotatebox{90}{Solar}}
 &96 & \boldres{.176} & \boldres{.227} & -- & -- & -- & -- & \secondres{.203} & \secondres{.237} & .234 & .286 & .290 & .378 \\
 &192 & \boldres{.192} & \boldres{.245} & -- & -- & -- & -- & \secondres{.233} & \secondres{.261} & .267 & .310 & .320 & .398 \\
 &336 & \boldres{.199} & \boldres{.250} & -- & -- & -- & -- & \secondres{.248} & \secondres{.273} & .290 & .315 & .353 & .415 \\
 &720 & \boldres{.203} & \boldres{.255} & -- & -- & -- & -- & \secondres{.249} & \secondres{.275} & .289 & .317 & .356 & .413 \\
\cmidrule(lr){2-14}
 &Avg & \boldres{.193} & \boldres{.244} & -- & -- & -- & -- & \secondres{.233} & \secondres{.262} & .270 & .307 & .330 & .401 \\
\midrule
\multirow{5}{*}{\rotatebox{90}{ECL}}
 &96 & \secondres{.130} & \secondres{.226} & .174 & .266 & .178 & .267 & .148 & .240 & \boldres{.129} & \boldres{.222} & .140 & .237 \\
 &192 & \boldres{.147} & \secondres{.241} & .182 & .273 & .187 & .274 & .162 & .253 & \secondres{.147} & \boldres{.240} & .153 & .249 \\
 &336 & \boldres{.162} & \boldres{.259} & .197 & .286 & .204 & .290 & .178 & .269 & \secondres{.163} & \secondres{.259} & .169 & .267 \\
 &720 & .206 & \secondres{.294} & .236 & .320 & .247 & .323 & .225 & .317 & \boldres{.197} & \boldres{.290} & \secondres{.203} & .301 \\
\cmidrule(lr){2-14}
 &Avg & \secondres{.161} & \secondres{.255} & .197 & .286 & .204 & .289 & .178 & .270 & \boldres{.159} & \boldres{.253} & .166 & .264 \\
\midrule
\multirow{5}{*}{\rotatebox{90}{Traffic}}
 &96 & \secondres{.379} & \secondres{.268} & -- & -- & .541 & .340 & .395 & .268 & \boldres{.360} & \boldres{.249} & .410 & .282 \\
 &192 & \secondres{.403} & .278 & -- & -- & .529 & .330 & .417 & \secondres{.276} & \boldres{.379} & \boldres{.256} & .423 & .287 \\
 &336 & \secondres{.415} & .284 & -- & -- & .545 & .334 & .433 & \secondres{.283} & \boldres{.392} & \boldres{.264} & .436 & .296 \\
 &720 & \secondres{.441} & \secondres{.299} & -- & -- & .580 & .351 & .467 & .302 & \boldres{.432} & \boldres{.286} & .466 & .315 \\
\cmidrule(lr){2-14}
 &Avg & \secondres{.410} & \secondres{.282} & -- & -- & .549 & .339 & .428 & .282 & \boldres{.391} & \boldres{.264} & .434 & .295 \\
\midrule
\multirow{5}{*}{\rotatebox{90}{ETTm2}}
 &96 & \boldres{.153} & \boldres{.254} & .174 & \secondres{.255} & .176 & .261 & .180 & .264 & \secondres{.166} & .256 & .167 & .260 \\
 &192 & \boldres{.195} & \boldres{.286} & .239 & .299 & .240 & .302 & .250 & .309 & \secondres{.223} & \secondres{.296} & .224 & .303 \\
 &336 & \boldres{.236} & \boldres{.317} & .301 & .340 & .299 & .342 & .311 & .348 & \secondres{.274} & \secondres{.329} & .281 & .342 \\
 &720 & \boldres{.289} & \boldres{.355} & .395 & .396 & .397 & .401 & .412 & .407 & \secondres{.362} & \secondres{.385} & .397 & .421 \\
\cmidrule(lr){2-14}
 &Avg & \boldres{.218} & \boldres{.303} & .277 & .323 & .278 & .327 & .288 & .332 & \secondres{.256} & \secondres{.317} & .267 & .332 \\
\midrule
\multirow{5}{*}{\rotatebox{90}{ETTh2}}
 &96 & \boldres{.235} & \boldres{.324} & .290 & .340 & .301 & .353 & .297 & .349 & \secondres{.274} & \secondres{.337} & .289 & .353 \\
 &192 & \boldres{.287} & \boldres{.361} & .375 & .392 & .379 & .405 & .380 & .400 & \secondres{.341} & \secondres{.382} & .383 & .418 \\
 &336 & \secondres{.333} & \secondres{.392} & .423 & .435 & .432 & .446 & .428 & .432 & \boldres{.329} & \boldres{.384} & .448 & .465 \\
 &720 & .429 & .450 & .443 & .449 & .446 & .463 & \secondres{.427} & \secondres{.445} & \boldres{.379} & \boldres{.422} & .605 & .551 \\
\cmidrule(lr){2-14}
 &Avg & \boldres{.321} & \secondres{.382} & .383 & .404 & .389 & .417 & .383 & .407 & \secondres{.331} & \boldres{.381} & .431 & .447 \\
\midrule
\multirow{5}{*}{\rotatebox{90}{ETTh1}}
 &96 & .447 & .450 & \boldres{.367} & \boldres{.395} & .374 & \secondres{.397} & .386 & .405 & \secondres{.370} & .400 & .375 & .399 \\
 &192 & .496 & .482 & .414 & \secondres{.420} & .419 & .429 & .441 & .436 & \secondres{.413} & .429 & \boldres{.405} & \boldres{.416} \\
 &336 & .584 & .534 & .445 & \boldres{.434} & .461 & .450 & .487 & .458 & \boldres{.422} & \secondres{.440} & \secondres{.439} & .443 \\
 &720 & .719 & .609 & \boldres{.444} & \boldres{.459} & .474 & \secondres{.467} & .503 & .491 & \secondres{.447} & .468 & .472 & .490 \\
\cmidrule(lr){2-14}
 &Avg & .562 & .519 & \secondres{.417} & \boldres{.427} & .432 & .436 & .454 & .447 & \boldres{.413} & \secondres{.434} & .423 & .437 \\
\midrule
\multirow{5}{*}{\rotatebox{90}{ETTm1}}
 &96 & .349 & .384 & .322 & .361 & .320 & .358 & .334 & .368 & \boldres{.293} & \secondres{.346} & \secondres{.299} & \boldres{.343} \\
 &192 & .392 & .411 & .357 & .383 & .364 & .383 & .377 & .391 & \boldres{.333} & \secondres{.370} & \secondres{.335} & \boldres{.365} \\
 &336 & .436 & .436 & .382 & .401 & .395 & .405 & .426 & .420 & \boldres{.369} & \secondres{.392} & \secondres{.369} & \boldres{.386} \\
 &720 & .495 & .474 & .445 & .435 & .457 & .440 & .491 & .459 & \boldres{.416} & \boldres{.420} & \secondres{.425} & \secondres{.421} \\
\cmidrule(lr){2-14}
 &Avg & .418 & .426 & .376 & .395 & .384 & .396 & .407 & .410 & \boldres{.353} & \secondres{.382} & \secondres{.357} & \boldres{.379} \\
\midrule
\multicolumn{2}{c}{1\textsuperscript{st} Count} & \textbf{15} & \textbf{15} & 2 & 3 & 0 & 0 & 0 & 0 & 14 & 10 & 1 & 4 \\
\bottomrule
\end{tabular}
\end{scriptsize}
\end{threeparttable}
}
\end{table}

Table~\ref{tab:main} presents results against selected published baselines. \textsc{DecompKAN} achieves the best or tied-best MSE on 15 of 32 dataset--horizon combinations among the baselines shown, though some margins are small (e.g., Weather H=96: .148 vs.\ PatchTST .149) and baseline seed variance is not reported, so not all margins may be statistically meaningful. Figure~\ref{fig:radar} visualizes the average MAE profile across all datasets under a controlled comparison. The results reveal a clear dataset-dependent pattern:

\textbf{Strong performance.}\quad \textsc{DecompKAN} achieves best MSE on Solar (4/4 horizons, $-17\%$ avg vs.\ iTransformer), ETTm2 (4/4), ECL (2/4; best at H=336, tied at H=192, loses H=96 and H=720 to PatchTST), Weather (3/4, approximately tied on average), and ETTh2 (2/4). These datasets share smooth temporal dynamics where decomposition and patching are effective; the advantage appears driven by temporal structure rather than cross-variate coupling, consistent with the channel-independent design.

\textbf{Weak performance.}\quad On ETTh1 (0/4), ETTm1 (0/4), and Traffic (0/4), \textsc{DecompKAN} underperforms published baselines. On ETTh1/ETTm1, TimeKAN and RMoK, which are also KAN-based architectures, outperform \textsc{DecompKAN}, so the gap is not simply KAN-vs-attention. On Traffic (862 variates), the underperformance cannot be attributed solely to channel independence, since PatchTST is also channel-independent and performs better; possible factors include the specific KAN core, decomposition kernel, or insufficient hyperparameter tuning. In preliminary basis-swap experiments (Appendix~\ref{app:synthetic}), replacing B-splines with RBF or Chebyshev did not close the ETTh1/ETTm1 gap, though these experiments were not exhaustive.

We note that this grouping is descriptive and post hoc. It reflects observed results rather than a principled a priori dataset taxonomy. A rigorous characterization of which dataset properties favor \textsc{DecompKAN} remains an open question.

\subsection{Controlled Comparison}

Table~\ref{tab:controlled} presents results where PatchTST, iTransformer, and DLinear are retrained with \textsc{DecompKAN}'s training protocol (same $L$, LR, batch size, epochs, cosine schedule, bidirectional augmentation, gradient clipping; Appendix~\ref{app:tuning}). This comparison should be interpreted with caution: the shared protocol was developed for \textsc{DecompKAN} and may disadvantage other architectures. Indeed, baselines often perform substantially worse than their published results under this protocol (e.g., PatchTST ETTh1 average MSE degrades from .413 published to .551 retrained; ETTm1 from .353 to .448). This indicates that the unified protocol is not equally suitable for all models.

\begin{table}[!htbp]
\caption{Controlled comparison---all models trained with \textsc{DecompKAN}'s recipe (Appendix~\ref{app:tuning}). Single-seed (seed=42). \boldres{Red}: best. \secondres{Blue}: second. First counts use unrounded values.}
\label{tab:controlled}
\vskip 0.02in
\centering
\begin{threeparttable}
\begin{scriptsize}
\renewcommand{\arraystretch}{0.78}
\setlength{\tabcolsep}{3.5pt}
\begin{tabular}{c|c|cc|cc|cc|cc}
\toprule
\multicolumn{2}{c}{\multirow{2}{*}{Models}} &
\multicolumn{2}{c}{\textbf{DecompKAN}} &
\multicolumn{2}{c}{PatchTST} &
\multicolumn{2}{c}{iTransformer} &
\multicolumn{2}{c}{DLinear} \\
\multicolumn{2}{c}{} &
\multicolumn{2}{c}{\textbf{(This work)}} &
\multicolumn{2}{c}{(same recipe)} &
\multicolumn{2}{c}{(same recipe)} &
\multicolumn{2}{c}{(same recipe)} \\
\cmidrule(lr){3-4}\cmidrule(lr){5-6}\cmidrule(lr){7-8}\cmidrule(lr){9-10}
\multicolumn{2}{c}{Metric} & MSE & MAE & MSE & MAE & MSE & MAE & MSE & MAE \\
\toprule
\multirow{5}{*}{\rotatebox{90}{Weather}}
& 96  & \boldres{0.146} & \boldres{0.194} & 0.176 & 0.232 & \secondres{0.168} & \secondres{0.215} & 0.181 & 0.245 \\
& 192  & \boldres{0.192} & \boldres{0.240} & 0.213 & 0.264 & \secondres{0.209} & \secondres{0.253} & 0.222 & 0.280 \\
& 336  & \boldres{0.244} & \boldres{0.279} & \secondres{0.260} & \secondres{0.296} & 0.265 & 0.297 & 0.280 & 0.337 \\
& 720  & \boldres{0.322} & \boldres{0.330} & 0.347 & 0.355 & \secondres{0.332} & \secondres{0.344} & 0.341 & 0.384 \\
\cmidrule(lr){2-10}
& Avg & \boldres{0.226} & \boldres{0.261} & 0.249 & 0.287 & \secondres{0.244} & \secondres{0.277} & 0.256 & 0.312 \\
\midrule
\multirow{5}{*}{\rotatebox{90}{Solar}}
& 96  & \boldres{0.176} & \boldres{0.227} & \secondres{0.179} & \secondres{0.236} & 0.187 & 0.236 & 0.221 & 0.298 \\
& 192  & \boldres{0.192} & \boldres{0.245} & \secondres{0.193} & \secondres{0.246} & 0.206 & 0.254 & 0.248 & 0.313 \\
& 336  & \secondres{0.199} & \boldres{0.250} & \boldres{0.197} & \secondres{0.254} & 0.213 & 0.265 & 0.269 & 0.327 \\
& 720  & \secondres{0.203} & \boldres{0.255} & \boldres{0.200} & \secondres{0.259} & 0.215 & 0.267 & 0.269 & 0.333 \\
\cmidrule(lr){2-10}
& Avg & \boldres{0.192} & \boldres{0.244} & \secondres{0.192} & \secondres{0.249} & 0.205 & 0.256 & 0.252 & 0.318 \\
\midrule
\multirow{5}{*}{\rotatebox{90}{ECL}}
& 96  & \boldres{0.130} & \boldres{0.226} & 0.149 & 0.258 & \secondres{0.132} & \secondres{0.228} & 0.135 & 0.233 \\
& 192  & \secondres{0.147} & \boldres{0.241} & 0.162 & 0.262 & 0.153 & 0.247 & \boldres{0.146} & \secondres{0.243} \\
& 336  & \boldres{0.162} & \boldres{0.259} & 0.181 & 0.286 & 0.166 & 0.263 & \secondres{0.162} & \secondres{0.261} \\
& 720  & 0.206 & \secondres{0.294} & 0.221 & 0.316 & \boldres{0.197} & \boldres{0.288} & \secondres{0.201} & 0.299 \\
\cmidrule(lr){2-10}
& Avg & \boldres{0.161} & \boldres{0.255} & 0.178 & 0.280 & 0.162 & \secondres{0.256} & \secondres{0.161} & 0.259 \\
\midrule
\multirow{5}{*}{\rotatebox{90}{Traffic}}
& 96  & \secondres{0.379} & \secondres{0.268} & 0.383 & 0.275 & \boldres{0.369} & \boldres{0.260} & 0.419 & 0.288 \\
& 192  & \secondres{0.403} & \secondres{0.278} & 0.406 & 0.288 & \boldres{0.389} & \boldres{0.270} & 0.433 & 0.294 \\
& 336  & 0.415 & 0.284 & \boldres{0.403} & \boldres{0.269} & \secondres{0.404} & \secondres{0.279} & 0.446 & 0.302 \\
& 720  & \secondres{0.441} & \secondres{0.299} & \boldres{0.433} & \boldres{0.288} & 0.447 & 0.310 & 0.471 & 0.319 \\
\cmidrule(lr){2-10}
& Avg & 0.410 & 0.282 & \secondres{0.406} & \boldres{0.280} & \boldres{0.402} & \secondres{0.280} & 0.442 & 0.301 \\
\midrule
\multirow{5}{*}{\rotatebox{90}{ETTh1}}
& 96  & \secondres{0.450} & \secondres{0.453} & 0.478 & 0.467 & 0.474 & 0.465 & \boldres{0.444} & \boldres{0.442} \\
& 192  & \secondres{0.495} & \secondres{0.481} & 0.512 & 0.493 & 0.500 & 0.483 & \boldres{0.489} & \boldres{0.471} \\
& 336  & 0.604 & 0.545 & 0.556 & 0.514 & \secondres{0.546} & \secondres{0.512} & \boldres{0.535} & \boldres{0.509} \\
& 720  & 0.729 & 0.613 & \secondres{0.657} & \secondres{0.579} & 0.657 & 0.583 & \boldres{0.632} & \boldres{0.573} \\
\cmidrule(lr){2-10}
& Avg & 0.570 & 0.523 & 0.551 & 0.513 & \secondres{0.544} & \secondres{0.511} & \boldres{0.525} & \boldres{0.499} \\
\midrule
\multirow{5}{*}{\rotatebox{90}{ETTh2}}
& 96  & \boldres{0.233} & \boldres{0.323} & 0.248 & 0.340 & 0.252 & 0.338 & \secondres{0.234} & \secondres{0.325} \\
& 192  & \boldres{0.289} & \boldres{0.362} & 0.310 & 0.382 & 0.307 & 0.372 & \secondres{0.293} & \secondres{0.369} \\
& 336  & \boldres{0.334} & \boldres{0.395} & 0.359 & 0.414 & 0.378 & 0.425 & \secondres{0.334} & \secondres{0.398} \\
& 720  & \boldres{0.428} & \boldres{0.450} & \secondres{0.457} & \secondres{0.460} & 0.464 & 0.471 & 0.671 & 0.583 \\
\cmidrule(lr){2-10}
& Avg & \boldres{0.321} & \boldres{0.382} & \secondres{0.344} & \secondres{0.399} & 0.350 & 0.401 & 0.383 & 0.419 \\
\midrule
\multirow{5}{*}{\rotatebox{90}{ETTm1}}
& 96  & \boldres{0.349} & \boldres{0.384} & 0.382 & 0.400 & 0.370 & 0.399 & \secondres{0.363} & \secondres{0.391} \\
& 192  & \boldres{0.392} & \boldres{0.411} & 0.417 & 0.428 & 0.430 & 0.431 & \secondres{0.404} & \secondres{0.414} \\
& 336  & \boldres{0.436} & \boldres{0.436} & 0.470 & 0.458 & 0.457 & 0.451 & \secondres{0.446} & \secondres{0.440} \\
& 720  & \boldres{0.495} & \secondres{0.474} & 0.524 & 0.497 & 0.518 & 0.485 & \secondres{0.496} & \boldres{0.468} \\
\cmidrule(lr){2-10}
& Avg & \boldres{0.418} & \boldres{0.426} & 0.448 & 0.446 & 0.444 & 0.442 & \secondres{0.427} & \secondres{0.428} \\
\midrule
\multirow{5}{*}{\rotatebox{90}{ETTm2}}
& 96  & \boldres{0.150} & \boldres{0.253} & 0.156 & 0.259 & 0.196 & 0.302 & \secondres{0.150} & \secondres{0.254} \\
& 192  & \secondres{0.195} & \boldres{0.286} & 0.200 & 0.295 & 0.221 & 0.316 & \boldres{0.188} & \secondres{0.286} \\
& 336  & \secondres{0.240} & \secondres{0.319} & 0.261 & 0.331 & 0.259 & 0.342 & \boldres{0.229} & \boldres{0.317} \\
& 720  & \boldres{0.300} & \boldres{0.360} & 0.336 & 0.396 & 0.319 & 0.379 & \secondres{0.312} & \secondres{0.377} \\
\cmidrule(lr){2-10}
& Avg & \secondres{0.221} & \boldres{0.304} & 0.238 & 0.320 & 0.249 & 0.335 & \boldres{0.220} & \secondres{0.308} \\
\midrule
\multirow{5}{*}{\rotatebox{90}{PPG}}
& 96  & \secondres{0.449} & \secondres{0.368} & \boldres{0.448} & \boldres{0.368} & 0.458 & 0.372 & 0.471 & 0.385 \\
& 192  & \boldres{0.494} & \boldres{0.396} & \secondres{0.496} & \secondres{0.398} & 0.508 & 0.402 & 0.517 & 0.413 \\
& 336  & \boldres{0.568} & \boldres{0.431} & \secondres{0.571} & \secondres{0.433} & 0.590 & 0.439 & 0.582 & 0.447 \\
& 720  & \secondres{0.760} & \secondres{0.502} & 0.781 & 0.510 & 0.767 & 0.503 & \boldres{0.727} & \boldres{0.507} \\
\cmidrule(lr){2-10}
& Avg & \boldres{0.568} & \boldres{0.424} & \secondres{0.574} & \secondres{0.427} & 0.581 & 0.429 & 0.574 & 0.438 \\
\midrule
\multicolumn{2}{c}{1\textsuperscript{st} Count} & \textbf{20} & \textbf{25} & 5 & 3 & 3 & 3 & 11 & 7 \\
\bottomrule
\end{tabular}
\end{scriptsize}
\end{threeparttable}
\end{table}

With the above caveats, the controlled comparison shows:

\textbf{Physics-structured datasets.}\quad Under the unified protocol (single seed=42), \textsc{DecompKAN} is competitive or superior on Weather (4/4), Solar (2/4 MSE; PatchTST is better at H=336 and H=720 by ${\sim}$0.002--0.003, though the average is approximately tied), and ECL (2/4 MSE, 3/4 MAE). The margins are modest, and since the unified protocol was originally developed for \textsc{DecompKAN} and multi-seed statistics are not available for the retrained baselines, this evidence is suggestive rather than conclusive.

\textbf{Protocol sensitivity.}\quad ETTm1 flips from 0/4 best results (published baselines) to 4/4 (retrained baselines). However, this reversal is likely driven by the unified protocol degrading baseline performance (PatchTST ETTm1 avg MSE: .353 published $\rightarrow$ .448 retrained) rather than by \textsc{DecompKAN} improving. TimeKAN and RMoK-B were not retrained. This result primarily illustrates the sensitivity of relative rankings to training protocol, not an architectural advantage.

\textbf{Cross-variate design choice.}
\label{sec:cross_variate}
During model development, several cross-variate architectures were explored informally (dense MLP mixing, GRU-based fusion, gated fusion with iTransformer, sparse GNN; see Appendix~\ref{app:cross_variate}). In preliminary experiments, none consistently improved average performance over the channel-independent baseline, leading to the adoption of channel independence in the final architecture. However, these explorations were not exhaustively tuned, and the comparison is qualitative rather than rigorous. A systematic study with equal hyperparameter budgets per variant would be needed to draw stronger conclusions about cross-variate modeling on these benchmarks.

\subsection{Ablation Study}

A component-level ablation (Appendix~\ref{app:ablation}) confirms that RevIN ($+$4.1--4.8\% degradation on Weather/ETTm2 when removed) and decomposition ($+$1.8--3.4\%) are the largest architectural contributors. Replacing KAN with linear or attention layers yields comparable or sometimes better results, demonstrating that the pipeline design matters more than the specific nonlinear layer. Bidirectional augmentation contributes substantially on ETTh1 ($+$10.0\% degradation when removed) but minimally on Weather ($+$0.6\%). The KAN formulation's primary advantage is inspectable latent edge functions rather than raw accuracy.

\subsection{Reproducibility}

To verify stability across random initializations, \textsc{DecompKAN} is evaluated with 10 random seeds on four datasets. Table~\ref{tab:seeds} reports mean $\pm$ standard deviation of MSE across seeds.

\begin{table}[!htbp]
\caption{Multi-seed evaluation (10 seeds). MSE reported as mean $\pm$ std.}
\label{tab:seeds}
\centering\small
\setlength{\tabcolsep}{4pt}
\begin{tabular}{c|cccc}
\toprule
$H$ & Weather & ETTh2 & ETTm2 & ETTh1 \\
\midrule
96  & .148$\pm$.001 & .235$\pm$.003 & .153$\pm$.003 & .447$\pm$.003 \\
192 & .192$\pm$.001 & .287$\pm$.002 & .195$\pm$.002 & .496$\pm$.001 \\
336 & .243$\pm$.001 & .333$\pm$.003 & .236$\pm$.006 & .584$\pm$.029 \\
720 & .321$\pm$.002 & .429$\pm$.001 & .289$\pm$.005 & .719$\pm$.025 \\
\bottomrule
\end{tabular}
\end{table}

Standard deviations are consistently small ($\leq 0.006$ for Weather, ETTh2, and ETTm2), confirming that \textsc{DecompKAN}'s results are stable across random initializations. ETTh1 shows slightly higher variance at long horizons ($H \geq 336$), consistent with the inherent difficulty of this behavioral dataset. Note that since multi-seed results are not available for the baselines, formal statistical significance tests (e.g., paired comparisons across seeds) cannot be conducted. The 10-seed means are used directly in Table~\ref{tab:main} for the four datasets evaluated here.

\section{Conclusion}

This work presented \textsc{DecompKAN}, a lightweight attention-free architecture (1.9--3.6M parameters) for long-term time series forecasting that combines trend--residual decomposition, channel-wise patching, learned normalization, and B-spline KAN edge functions. The architecture provides competitive accuracy alongside inspectable learned nonlinear functions. Each KAN edge function is a directly visualizable 1D transformation over learned patch-embedding coordinates, offering a form of architectural transparency that complements but differs from attention-weight or gradient-based inspection in other architectures.

On standard benchmarks, \textsc{DecompKAN} achieves best or tied-best MSE on 15/32 dataset--horizon combinations among selected published baselines, with particular strength on datasets with smooth, decomposable temporal structure (Solar $-17\%$, ECL $-10\%$ vs.\ iTransformer; Weather approximately tied). The model underperforms on datasets where cross-variate dependencies or complex regime changes dominate (ETTh1, ETTm1, Traffic). Visualization of learned edge functions reveals qualitatively different latent nonlinearities across domains, though mapping these to raw physical variables requires further attribution analysis.

\textbf{Limitations.}\quad The channel-independent design cannot capture cross-variate dependencies that are stable and causal. On datasets with complex regime changes with complex regime changes (ETTh1, ETTm1), \textsc{DecompKAN} underperforms other architectures, including other KAN-based models (TimeKAN, RMoK), suggesting that the specific design choices (dual-branch decomposition, channel independence) are not optimal for all data types. Table~\ref{tab:main} uses 10-seed means for four datasets and single-seed results for the remaining four, while baselines are published numbers with unclear seed protocols. This mixed reporting and the absence of multi-seed baselines limit the statistical strength of small-margin wins.

\textbf{Reproducibility.}\quad The model is implemented in PyTorch using the \texttt{efficient-kan} library for B-spline KAN layers. All experiments use fixed random seeds. Code, trained models, and per-seed results will be released upon publication.

\textbf{Future work.}\quad Three directions are promising: (1)~a rigorous study of alternative basis functions (Fourier, Chebyshev, RBF, Gabor) within the decomposed Patch-KAN framework, as preliminary synthetic experiments (Appendix~\ref{app:synthetic}) suggest that basis choice interacts with signal structure in ways that merit deeper investigation; (2)~application to atmospheric and climate datasets where position-dependent physics is dominant; (3)~sparse graph neural network integration for cross-variate modeling in datasets with known causal structure, building on preliminary evidence that learned sparse adjacency can discover physically meaningful variate relationships.

\bibliographystyle{plainnat}
\bibliography{references}

@inproceedings{zhou2021informer,
  title={Informer: Beyond Efficient Transformer for Long Sequence Time-Series Forecasting},
  author={Zhou, Haoyi and Zhang, Shanghang and Peng, Jieqi and Zhang, Shuai and Li, Jianxin and Xiong, Hui and Zhang, Wancai},
  booktitle={AAAI},
  year={2021}
}

@inproceedings{wu2021autoformer,
  title={Autoformer: Decomposition Transformers with Auto-Correlation for Long-Term Series Forecasting},
  author={Wu, Haixu and Xu, Jiehui and Wang, Jianmin and Long, Mingsheng},
  booktitle={NeurIPS},
  year={2021}
}

@inproceedings{zhou2022fedformer,
  title={{FEDformer}: Frequency Enhanced Decomposed Transformer for Long-term Series Forecasting},
  author={Zhou, Tian and Ma, Ziqing and Wen, Qingsong and Wang, Xue and Sun, Liang and Jin, Rong},
  booktitle={ICML},
  year={2022}
}

@inproceedings{zeng2023dlinear,
  title={Are Transformers Effective for Time Series Forecasting?},
  author={Zeng, Ailing and Chen, Muxi and Zhang, Lei and Xu, Qiang},
  booktitle={AAAI},
  year={2023}
}

@inproceedings{nie2023patchtst,
  title={A Time Series is Worth 64 Words: Long-term Forecasting with Transformers},
  author={Nie, Yuqi and Nguyen, Nam H and Sinthong, Phanwadee and Kalagnanam, Jayant},
  booktitle={ICLR},
  year={2023}
}

@inproceedings{liu2024itransformer,
  title={{iTransformer}: Inverted Transformers Are Effective for Time Series Forecasting},
  author={Liu, Yong and Hu, Tengge and Zhang, Haoran and Wu, Haixu and Wang, Shiyu and Ma, Lintao and Long, Mingsheng},
  booktitle={ICLR},
  year={2024}
}

@inproceedings{wu2023timesnet,
  title={{TimesNet}: Temporal 2D-Variation Modeling for General Time Series Analysis},
  author={Wu, Haixu and Hu, Tengge and Liu, Yong and Zhou, Hang and Wang, Jianmin and Long, Mingsheng},
  booktitle={ICLR},
  year={2023}
}

@inproceedings{kim2022revin,
  title={Reversible Instance Normalization for Accurate Time-Series Forecasting against Distribution Shift},
  author={Kim, Taesung and Kim, Jinhee and Tae, Yunwon and Park, Cheonbok and Choi, Jang-Ho and Choo, Jaegul},
  booktitle={ICLR},
  year={2022}
}

@inproceedings{liu2022nonstationary,
  title={Non-stationary Transformers: Exploring the Stationarity in Time Series Forecasting},
  author={Liu, Yong and Wu, Haixu and Wang, Jianmin and Long, Mingsheng},
  booktitle={NeurIPS},
  year={2022}
}

@article{fan2023dishts,
  title={{Dish-TS}: A General Paradigm for Alleviating Distribution Shift in Time Series Forecasting},
  author={Fan, Wei and Wang, Pengyang and Wang, Dongkun and Wang, Dongjie and Zhou, Yuanchun and Fu, Yanjie},
  journal={AAAI},
  year={2023}
}

@inproceedings{liu2024kan,
  title={{KAN}: Kolmogorov-Arnold Networks},
  author={Liu, Ziming and Wang, Yixuan and Vaidya, Sachin and Ruehle, Fabian and Halverson, James and Solja{\v{c}}i{\'c}, Marin and Hou, Thomas Y and Tegmark, Max},
  booktitle={ICLR},
  year={2025}
}

@article{xu2024kan4tsf,
  title={{KAN4TSF}: Are {KAN} and {KAN}-based models Effective for Time Series Forecasting?},
  author={Han, Xiao and Zhang, Xinfeng and Wu, Yiling and Zhang, Zhenduo and Wu, Zhe},
  journal={arXiv preprint arXiv:2408.11306},
  year={2024}
}

@article{qiu2024timekan,
  title={{TimeKAN}: {KAN}-based Frequency Decomposition Learning Architecture for Long-term Time Series Forecasting},
  author={Huang, Songtao and Zhao, Zhen and Li, Can and Bai, Lei},
  journal={arXiv preprint arXiv:2502.06910},
  year={2025}
}

@inproceedings{woo2024moirai,
  title={Unified Training of Universal Time Series Forecasting Transformers},
  author={Woo, Gerald and Liu, Chenghao and Kumar, Akshat and Xiong, Caiming and Savarese, Silvio and Sahoo, Doyen},
  booktitle={ICML},
  year={2024}
}

@article{ansari2024chronos,
  title={Chronos: Learning the Language of Time Series},
  author={Ansari, Abdul Fatir and Stella, Lorenzo and Turkmen, Caner and Zhang, Xiyuan and Mercado, Pedro and Shen, Huibin and Shchur, Oleksandr and Rangapuram, Syama Sundar and Arango, Sebastian Pineda and Kapoor, Shubham and others},
  journal={Transactions on Machine Learning Research (TMLR)},
  year={2024}
}

@article{das2024timesfm,
  title={A Decoder-Only Foundation Model for Time-Series Forecasting},
  author={Das, Abhimanyu and Kong, Weihao and Sen, Rajat and Zhou, Yichen},
  journal={ICML},
  year={2024}
}

@article{jin2024timellm,
  title={{Time-LLM}: Time Series Forecasting by Reprogramming Large Language Models},
  author={Jin, Ming and Wang, Shiyu and Ma, Lintao and Chu, Zhixuan and Zhang, James Y. and Shi, Xiaoming and Chen, Pin-Yu and Liang, Yuxuan and Li, Yuan-Fang and Pan, Shirui and Wen, Qingsong},
  journal={ICLR},
  year={2024}
}

@inproceedings{vaswani2017attention,
  title={Attention Is All You Need},
  author={Vaswani, Ashish and Shazeer, Noam and Parmar, Niki and Uszkoreit, Jakob and Jones, Llion and Gomez, Aidan N and Kaiser, {\L}ukasz and Polosukhin, Illia},
  booktitle={NeurIPS},
  year={2017}
}

@article{kolmogorov1957representation,
  title={On the Representation of Continuous Functions of Several Variables by Superposition of Continuous Functions of One Variable and Addition},
  author={Kolmogorov, Andrey N},
  journal={Doklady Akademii Nauk SSSR},
  volume={114},
  pages={953--956},
  year={1957}
}

@book{deboor1978splines,
  title={A Practical Guide to Splines},
  author={de Boor, Carl},
  publisher={Springer},
  year={1978}
}

@inproceedings{lai2018lstnet,
  title={Modeling Long- and Short-Term Temporal Patterns with Deep Neural Networks},
  author={Lai, Guokun and Chang, Wei-Cheng and Yang, Yiming and Liu, Hanxiao},
  booktitle={SIGIR},
  year={2018}
}

@inproceedings{velickovic2018gat,
  title={Graph Attention Networks},
  author={Veli{\v{c}}kovi{\'c}, Petar and Cucurull, Guillem and Casanova, Arantxa and Romero, Adriana and Li{\`o}, Pietro and Bengio, Yoshua},
  booktitle={ICLR},
  year={2018}
}

@inproceedings{cho2014gru,
  title={Learning Phrase Representations using {RNN} Encoder--Decoder for Statistical Machine Translation},
  author={Cho, Kyunghyun and van Merri{\"e}nboer, Bart and Gulcehre, Caglar and Bahdanau, Dzmitry and Bougares, Fethi and Schwenk, Holger and Bengio, Yoshua},
  booktitle={EMNLP},
  year={2014}
}

@article{kingma2015adam,
  title={Adam: A Method for Stochastic Optimization},
  author={Kingma, Diederik P and Ba, Jimmy},
  journal={ICLR},
  year={2015}
}

@inproceedings{chen2025simpletm,
  title={{SimpleTM}: A Simple Baseline for Multivariate Time Series Forecasting},
  author={Chen, Hui and Luong, Viet and Mukherjee, Lopamudra and Singh, Vikas},
  booktitle={ICLR},
  year={2025}
}

@misc{reiss2019ppg,
  title={{PPG-DaLiA}: A {PPG} Dataset for Motion Compensation and Heart Rate Estimation in Daily Life Activities},
  author={Reiss, Attila and Indlekofer, Ina and Schmidt, Philip and Van Laerhoven, Kristof},
  year={2019},
  note={UCI Machine Learning Repository. \url{https://doi.org/10.24432/C53890}},
}

@inproceedings{lu2024cats,
  title={{CATS}: Enhancing Multivariate Time Series Forecasting by Constructing Auxiliary Time Series as Exogenous Variables},
  author={Lu, Jiecheng and Han, Xu and Sun, Yan and Yang, Shihao},
  booktitle={ICML},
  year={2024}
}

\appendix

\section{Per-Dataset Tuning Details}
\label{app:tuning}

Best hyperparameters per dataset from the tuning sweep (12 configs, selected by validation MSE at $H{=}96$):

\begin{table}[h]
\caption{Per-dataset hyperparameters. All other parameters are fixed: $P{=}16$, $S{=}8$, $d{=}32$, KAN hidden$={64}$, depth$={2}$, $G{=}5$, spline order $p{=}3$, MA kernel $K_{\mathrm{MA}}{=}25$.}
\centering\small
\begin{tabular}{lccc}
\toprule
Dataset & Learning Rate & Bidirectional & Lookback $L$ \\
\midrule
Weather & $10^{-3}$ & On & 336 \\
Solar & $2 \times 10^{-4}$ & On & 336 \\
ECL & $5 \times 10^{-4}$ & On & 512 \\
Traffic & $5 \times 10^{-4}$ & On & 336 \\
ETTh2 & $10^{-3}$ & On & 336 \\
ETTh1 & $2 \times 10^{-4}$ & On & 336 \\
ETTm2 & $10^{-3}$ & Off & 336 \\
ETTm1 & $2 \times 10^{-4}$ & On & 512 \\
PPG-DaLiA & $5 \times 10^{-4}$ & Off & 336 \\
\bottomrule
\end{tabular}
\end{table}

\section{Parameter Counts}
\label{tab:params}

The parameter count depends on lookback length $L$ (which determines the number of patches $N$) and forecast horizon $H$ (which determines the KAN output dimension).

\begin{table}[h]
\caption{Parameter counts by lookback length and forecast horizon. The difference is driven by the KAN input dimension ($Nd$ where $N = \lfloor(L{-}P)/S\rfloor{+}1$) and output dimension ($H$).}
\centering\small
\begin{tabular}{c|cccc}
\toprule
& $H{=}96$ & $H{=}192$ & $H{=}336$ & $H{=}720$ \\
\midrule
$L{=}336$ & 1.90M & 2.02M & 2.20M & 2.70M \\
$L{=}512$ & 2.80M & 2.93M & 3.11M & 3.60M \\
\bottomrule
\end{tabular}
\end{table}

\section{Ablation Study}
\label{app:ablation}

Table~\ref{tab:ablation} validates each component on three representative datasets at $H{=}96$.

\begin{table}[!htbp]
\caption{Ablation study (MSE, $H{=}96$, single seed=42; values differ from 10-seed means in Table~\ref{tab:main}). Each row removes or replaces one component from the full \textsc{DecompKAN}.}
\label{tab:ablation}
\centering\small
\begin{tabular}{lccc}
\toprule
Configuration & Weather & ETTm2 & ETTh1 \\
\midrule
\textsc{DecompKAN} (full) & \textbf{.146} & .150 & .450 \\
\quad $-$ decomposition & .148 (+1.8\%) & .155 (+3.4\%) & .451 (+0.3\%) \\
\quad $-$ adaptive norm & .147 (+0.5\%) & .151 (+0.4\%) & .446 ($-$1.0\%) \\
\quad $-$ RevIN & .152 (+4.1\%) & .157 (+4.8\%) & .454 (+0.9\%) \\
\quad KAN $\rightarrow$ Linear & .150 (+2.5\%) & \textbf{.148} ($-$1.4\%) & \textbf{.438} ($-$2.8\%) \\
\quad KAN $\rightarrow$ Attention & .147 (+1.0\%) & .148 ($-$1.5\%) & .445 ($-$1.1\%) \\
\bottomrule
\end{tabular}
\end{table}

\textbf{RevIN} is the most impactful single component: removing it degrades all three datasets (Weather $+$4.1\%, ETTm2 $+$4.8\%, ETTh1 $+$0.9\%).

\textbf{Decomposition} provides consistent improvement on Weather ($+$1.8\%) and ETTm2 ($+$3.4\%), with minimal effect on ETTh1 ($+$0.3\%).

\textbf{KAN vs.\ alternatives.}\quad Replacing KAN with linear layers degrades Weather modestly ($+$2.5\%) but actually \emph{improves} ETTm2 ($-$1.4\%) and ETTh1 ($-$2.8\%). Replacing KAN with attention shows a similar pattern: marginally worse on Weather ($+$1.0\%) but better on ETTm2 ($-$1.5\%) and ETTh1 ($-$1.1\%). This demonstrates that within the pipeline, the choice of KAN over attention or linear layers provides only a small, dataset-dependent benefit. The primary value of \textsc{DecompKAN} lies in the full pipeline design, rather than in any single component.

\textbf{Adaptive normalization} has minimal impact (0.4--1.0\% across datasets), suggesting RevIN alone captures most of the normalization benefit.

\begin{figure}[h]
\centering
\includegraphics[width=\textwidth]{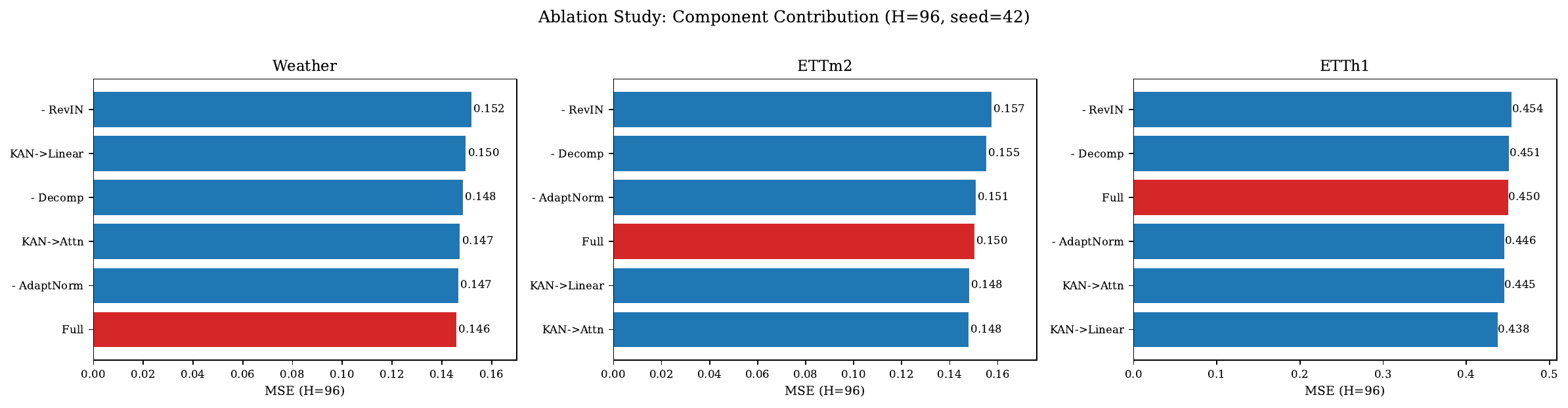}
\caption{Ablation study at $H{=}96$ (seed=42). Bars are sorted by MSE within each dataset. Full \textsc{DecompKAN} (red) is best only on Weather; on ETTm2 and ETTh1, replacing KAN with linear or attention layers yields equal or lower MSE, illustrating that the pipeline design matters more than the KAN layer.}
\label{fig:ablation}
\end{figure}

\textbf{Bidirectional augmentation.}\quad Table~\ref{tab:bidir_ablation} isolates the effect of bidirectional augmentation, which doubles the effective training set by adding time-reversed windows. On ETTh1, removing augmentation degrades MSE by $+$10.0\%, confirming it is a substantial contributor on that dataset. On Weather, the effect is small ($+$0.6\%). This clarifies that the reported ETTh1 performance depends partly on augmentation, while Weather performance is driven primarily by the architecture.

\begin{table}[h]
\centering\small
\caption{Bidirectional augmentation ablation ($H{=}96$, seed=42). Note: full-model values differ slightly from Table~\ref{tab:ablation} (0.1471 vs.\ 0.146) due to a reimplementation of the windowing function between runs; relative degradations are consistent.}
\label{tab:bidir_ablation}
\begin{tabular}{lccc}
\toprule
Dataset & With bidir (MSE) & Without bidir (MSE) & Degradation \\
\midrule
Weather & 0.1471 & 0.1480 & $+$0.6\% \\
ETTh1 & 0.4525 & 0.4975 & $+$10.0\% \\
\bottomrule
\end{tabular}
\end{table}

\section{Synthetic Experiments: Design Motivation}
\label{app:synthetic}

The \textsc{DecompKAN} architecture emerged from a series of synthetic experiments that progressively revealed the strengths and limitations of B-spline KAN on controlled signals. These experiments should be interpreted as design motivation rather than rigorous claims; each finding directly informed a specific architectural choice.

\textbf{Step 1: Where B-splines excel and fail.}\quad The investigation began with simple univariate signals ($L{=}96 \rightarrow H{=}96$). On bounded periodic signals (sine, sine+triangular, sine+cosine mixtures), B-spline KAN outperformed MLP by 6--37$\times$ on MSE: local polynomial pieces naturally represent piecewise-smooth waveforms. However, adding a linear trend ($0.002t$) completely reversed the result: MLP won 9/10 random trials by 2--14$\times$. The cause is that B-splines are defined on a fixed grid, and test values that exceed the training range force extrapolation beyond knot boundaries where polynomials drift. \emph{Design consequence:} input normalization (RevIN + adaptive) must remove trends before B-splines see the data.

\textbf{Step 2: Decomposition as the solution.}\quad A hybrid model was tested with moving-average decomposition into trend and residual, with MLP forecasting the trend and KAN forecasting the residual. On the same sine+cosine+trend signals where pure KAN failed, the hybrid won 9/10 trials, achieving 10$\times$ lower MSE than MLP and 76$\times$ lower than pure KAN (excluding one degenerate trial). Each branch specializes: MLP extrapolates the linear trend naturally; KAN handles the bounded residual in its sweet spot. \emph{Design consequence:} the dual-branch trend--residual architecture.

\textbf{Step 3: Capacity limits and patching.}\quad Signal complexity was increased by summing $k$ random sinusoids ($k{=}1$ to 10). B-spline KAN dominated at $k \leq 2$ but collapsed at $k \geq 3$, regardless of width (64 to 512 hidden) or grid resolution (5 to 20). Even a 2.4M-parameter KAN lost to a 49K MLP at $k{=}3$. The bottleneck is architectural: each B-spline edge applies a 1D nonlinearity to one scalar input, so the hidden layer must disentangle overlapping frequencies from a summed signal, which B-spline composition does inefficiently. Patch embedding partially mitigated this, improving KAN by 9--22$\times$ at $k{=}3$--$9$ and pushing the crossover from $k{=}3$ to $k{=}4$: converting raw time values into learned local representations reduces the frequency-disentanglement burden. \emph{Design consequence:} channel-wise patching before the KAN layers.

\textbf{Step 4: Basis function comparison.}\quad Eight basis functions (B-spline, Fourier, Chebyshev, RBF, Gabor, Haar wavelet, Morlet wavelet, MLP) were compared on both synthetic and real data. On synthetic signals, the choice of basis produced dramatic differences: time-aware Gabor and Fourier bases achieved near-perfect MSE on stationary periodic signals, while Chebyshev dominated on trending signals. However, on real datasets (Weather, ECL, ETTh1), all bases performed within a 5--10\% range of each other. When B-spline, RBF, and Chebyshev were swapped into the full \textsc{DecompKAN} architecture across five datasets, no basis closed the ETTh1/ETTm1 performance gap vs.\ published SOTA. \emph{Design consequence:} the pipeline (decomposition, patching, normalization) matters more than the specific nonlinear basis, which is the central finding of the ablation study (Table~\ref{tab:ablation}).

\textbf{Summary.}\quad Each experiment exposed a specific failure mode and motivated a corresponding design choice: normalization to handle trends (Step~1), decomposition to let branches specialize (Step~2), patching to reduce the frequency-disentanglement burden (Step~3), and the broader finding that pipeline design dominates basis selection (Step~4). These synthetic results are preliminary and not exhaustive, but they provide concrete motivation for the architectural decisions in \textsc{DecompKAN}.

\section{Cross-Variate Exploration}
\label{app:cross_variate}

\textsc{DecompKAN} uses a channel-independent design where all variates share the same weights. This choice followed a systematic exploration of cross-variate architectures during model development (10 variants tested across 4 datasets at $H{=}96$). The consistent finding was that more learnable cross-variate parameters led to more overfitting, and no approach improved average performance across all datasets.

\textbf{Dense MLP mixing.}\quad A fully-connected layer across variates after the KAN ($O(C^2)$ parameters). Overfitted on datasets with spurious inter-variate correlations (e.g., Exchange FX rates, which co-move due to external factors like monetary policy rather than direct physical coupling).

\textbf{Attention-based mixing.}\quad Bottleneck attention (13K--170K params), gated attention, and semantic-identity attention~\citep{velickovic2018gat} were tested. Attention helped on Weather (3/4 horizons), where physical laws create stable causal links between temperature, humidity, and pressure, but degraded Exchange and was neutral on ETT. The best attention variant used GPT-derived semantic embeddings of variate names as identity tokens, providing prior knowledge about which variates are related without learning it from data.

\textbf{Fixed semantic mixing.}\quad The most successful approach used a fixed similarity matrix derived from GPT embeddings of variate descriptions, with a single learnable scalar $\beta$ controlling mixing strength. With only 1 learnable cross-variate parameter, it could not overfit. The learned $\beta$ values were revealing: Weather $\beta{=}0.027$ (favors mixing; physics-based coupling), Exchange $\beta{=}0.010$ (near zero; co-movement driven by external factors).

\textbf{GRU sequence fusion.}\quad GRU~\citep{cho2014gru} encoder before cross-variate attention. Marginal gains on ETTh1 but degraded Weather and Solar, confirming that sequential cross-variate processing adds complexity without consistent benefit.

\textbf{Sparse GNN (GATConv).}\quad Learned node embeddings with sparse adjacency (top-$k$ neighbors per variate)~\citep{velickovic2018gat}. Graph attention discovered physically meaningful structure (e.g., thermal coupling among ETT transformer sensors) but did not improve benchmark performance, suggesting that discoverable structure in these datasets is insufficient to overcome the regularization benefit of channel independence.

\textbf{Conclusion.}\quad In preliminary experiments, cross-variate mixing appeared helpful only on datasets with stable, physically interpretable inter-variable relationships (e.g., Weather). On most standard benchmarks, inter-variate correlations may arise from external factors rather than direct coupling, and channel independence provides stronger implicit regularization. This led to the adoption of channel-independent design in \textsc{DecompKAN}, though this limits the model on datasets with genuine cross-variate dependencies (e.g., Traffic). These observations are qualitative and were not tested with formal causal methods.

\section{Interpretability: Visualizing Learned Edge Functions}
\label{app:interpretability}
\enlargethispage{2\baselineskip}

A distinguishing property of KAN-based architectures is that every edge function $\phi_{i \to j}: \mathbb{R} \to \mathbb{R}$ is an explicit, inspectable 1D transformation. Unlike attention weights (which indicate \emph{where} a model attends but not \emph{what computation} it performs) or gradient saliency maps (local linear approximations computed post-hoc), KAN edge functions are the actual learned nonlinearities through which data flows. This section visualizes edge functions learned on two datasets with distinct temporal characteristics: Weather (21 meteorological variates) and PPG-DaLiA~\citep{reiss2019ppg} (15 sensor channels: ECG, respiration, BVP, EDA, temperature, and 3-axis accelerometry from chest and wrist devices, resampled to 1Hz).

\begin{figure}[h]
\centering
\includegraphics[width=\textwidth]{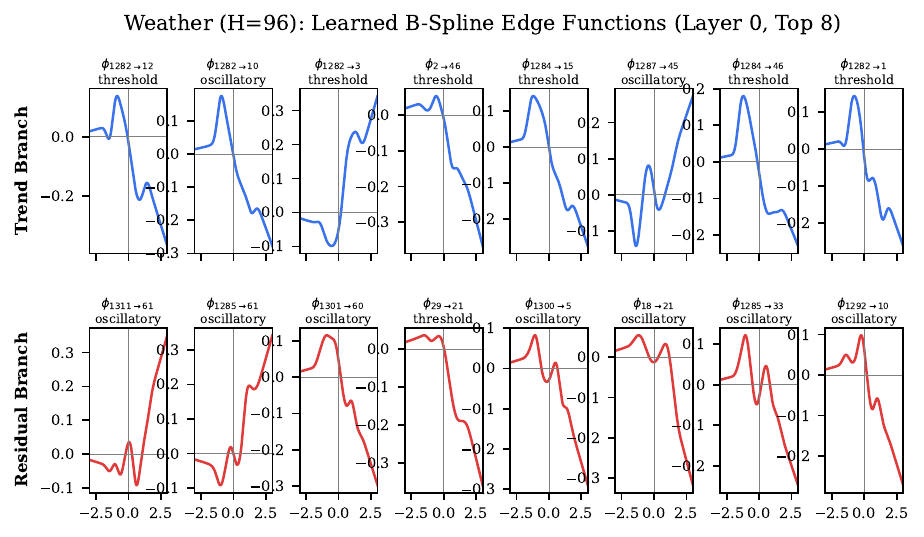}
\caption{Learned B-spline edge functions $\phi_{i \to j}(x)$ from the first KAN layer of \textsc{DecompKAN} trained on Weather ($H{=}96$). Top row: trend branch (blue); bottom row: residual branch (red). Each subplot shows one edge's activation function over $x \in [-3, 3]$. Edge labels indicate shape classification. The trend branch learns smoother functions; the residual branch learns sharper, more oscillatory transformations.}
\label{fig:edge_functions}
\end{figure}

\textbf{Qualitative analysis of edge functions.}\quad Figure~\ref{fig:edge_functions} displays the top 8 most active edge functions (ranked by activation range $\max \phi - \min \phi$) from the first KAN layer of each branch. Two observations emerge:

\begin{enumerate}[nosep]
    \item \textbf{Branch specialization.} The trend branch learns predominantly smooth, slowly varying functions (gradual slopes, soft thresholds), consistent with its role in capturing low-frequency dynamics. The residual branch learns sharper, more oscillatory functions with steeper gradients, appropriate for modeling higher-frequency seasonal and irregular components.
    \item \textbf{Functional diversity.} Even within a single layer, edges learn qualitatively different shapes: smooth monotone mappings, sharp threshold transitions, oscillatory patterns, and near-identity functions. This diversity arises naturally from training without any explicit regularization on edge function shape, suggesting that B-spline KAN layers discover a heterogeneous set of nonlinear features.
\end{enumerate}

\begin{figure}[h]
\centering
\includegraphics[width=\textwidth]{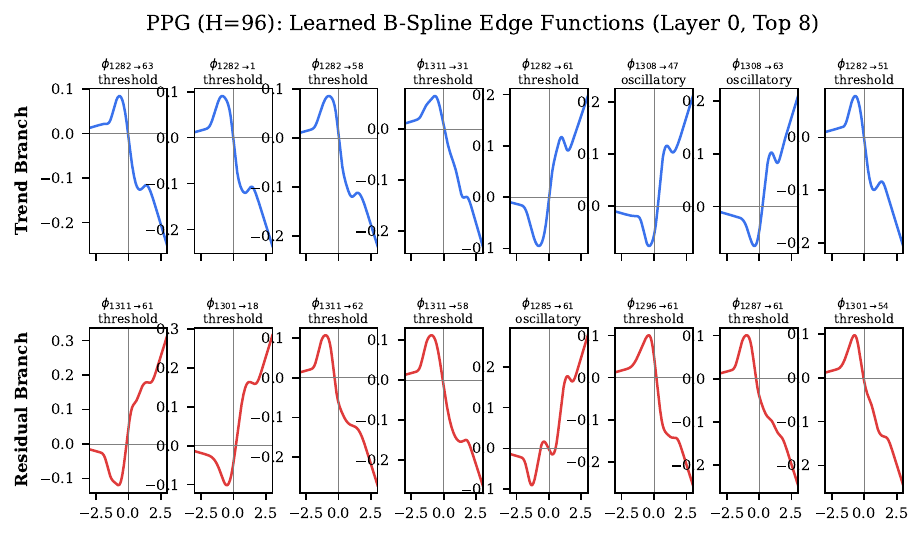}
\caption{Learned B-spline edge functions on PPG-DaLiA ($H{=}96$, 15 sensor channels). Unlike Weather (Figure~\ref{fig:edge_functions}), the PPG model learns predominantly threshold-like latent edge functions.}
\label{fig:edge_ppg}
\end{figure}

\textbf{Cross-dataset edge patterns.}\quad Figure~\ref{fig:edge_ppg} shows the same analysis on PPG-DaLiA (15 sensor channels from chest and wrist devices). Compared to Weather (Figure~\ref{fig:edge_functions}), the PPG model learns markedly different edge shapes: predominantly threshold-like functions rather than oscillatory patterns. The trend branch learns smoother transitions, while the residual branch learns sharper step-like functions. This qualitative difference emerges without any domain-specific configuration, though we note these functions operate on latent patch-embedding coordinates and mapping them to raw sensor channels requires additional attribution analysis.

\textbf{PPG-DaLiA results.}\quad Full controlled-comparison results across all four horizons are reported in Table~\ref{tab:controlled}. \textsc{DecompKAN} ranks first on 2 of 4 horizons (H=192, H=336) and is within 0.001 MSE of PatchTST at H=96, consistent with the Weather pattern where physics-coupled datasets favor channel-independent architectures with strong temporal processing.

\begin{figure}[h]
\centering
\includegraphics[width=0.85\textwidth]{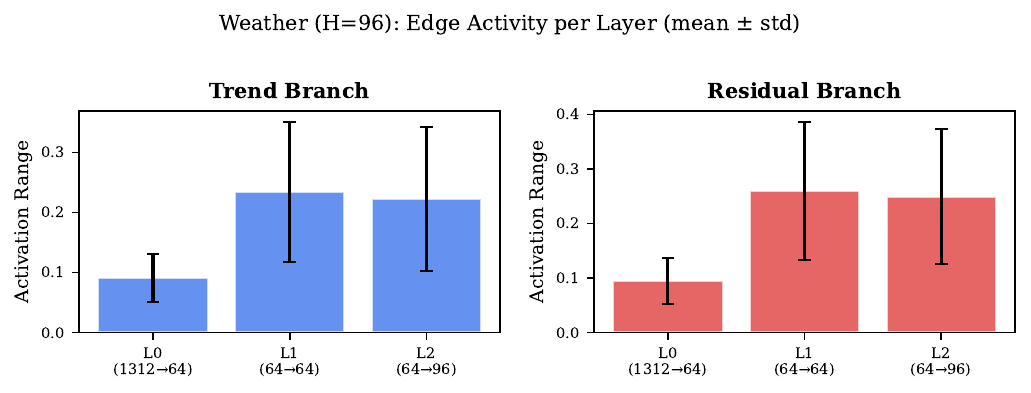}
\caption{Mean activation range ($\max \phi - \min \phi$) per KAN layer on Weather, with error bars showing $\pm 1$ standard deviation across all edges. Layer 0 ($1312 \to 64$) exhibits structured sparsity with low mean range, while deeper layers are uniformly more active.}
\label{fig:layer_activity}
\end{figure}

\textbf{Structured sparsity.}\quad Figure~\ref{fig:layer_activity} shows the mean activation range per layer across both branches on Weather. The first layer ($1312 \to 64$, mapping flattened patch embeddings to hidden features) has substantially lower mean activation range (${\sim}0.09$) than subsequent layers (${\sim}0.23$--$0.26$), indicating that most edges in this wide input layer learn near-zero functions. This structured sparsity means the first KAN layer acts as a learned feature selector: out of 83{,}968 edges, only a fraction carry significant signal. The deeper layers ($64 \to 64$ and $64 \to 96$) are uniformly more active, suggesting that the KAN concentrates its representational capacity in the hidden-to-output mapping rather than the input projection. The same pattern holds on PPG (L0 mean range ${\sim}0.07$, L1--L2 ${\sim}0.22$).

\textbf{Implications for safety-critical applications.}\quad In medical monitoring, climate science, and energy management, regulatory frameworks increasingly require transparency beyond accuracy metrics. Unlike attention maps (which show token relevance but not functional transformations) or MLP activations (entangled across neurons), KAN edge functions provide a faithful per-edge representation of learned computation. Domain experts can inspect any $\phi_{i \to j}$ to validate physical plausibility or flag failure modes. That Weather and PPG models learn qualitatively different edge shapes (oscillatory vs.\ threshold) without domain-specific configuration demonstrates that B-spline KAN layers adapt their functional vocabulary to the data.

\end{document}